%% file: main.tex
\documentclass[10pt,twocolumn,letterpaper]{article}

\usepackage{iccv}
\usepackage{times}
\usepackage{epsfig}
\usepackage{graphicx}
\usepackage{amsmath}
\usepackage{amssymb}
\usepackage{booktabs}
\usepackage{color}
\usepackage{multirow}
\usepackage{subfig}
\usepackage{caption}
\usepackage{mmstyle}
\usepackage{mathtools}

\usepackage[pagebackref=true,breaklinks=true,letterpaper=true,colorlinks,bookmarks=false]{hyperref}

\iccvfinalcopy 

\def\iccvPaperID{3035} 

\ificcvfinal\fi
\begin{document}

\title{Recursive Visual Sound Separation Using Minus-Plus Net}

\author{Xudong Xu~~~~~~\qquad\qquad~~~~~~Bo Dai~~~~~~\qquad\qquad~~~~~~Dahua Lin\\
CUHK-SenseTime Joint Lab, The Chinese University of Hong Kong\\
\texttt{xx018@ie.cuhk.edu.hk}~~~~~\texttt{bdai@ie.cuhk.edu.hk}~~~~~\texttt{dhlin@ie.cuhk.edu.hk}
}

\maketitle
\ificcvfinal\thispagestyle{empty}\fi

\input{abstract.tex}
\input{intro.tex}

\input{relwork.tex}

\input{frmwork.tex}

\input{exprt.tex}

\input{concls.tex}

{\small
\bibliographystyle{ieee_fullname}
\bibliography{main}
}

\end{document}

%% file: abstract.tex
\begin{abstract}
\vspace{-5pt}
Sounds provide rich semantics, complementary to visual data,
for many tasks.
However,
in practice,
sounds from multiple sources are often mixed together.
In this paper we propose a novel framework,
referred to as MinusPlus Network (MP-Net),
for the task of visual sound separation.
MP-Net separates sounds recursively in the order of average energy
\footnote{In this paper, average energy of sound stands for the average energy of its spectrogram.},
removing the separated sound from the mixture at the end of each prediction,
until the mixture becomes empty or contains only noise.
In this way,
MP-Net could be applied to sound mixtures with arbitrary numbers and types of sounds.
Moreover,
while MP-Net keeps removing sounds with large energy from the mixture,
sounds with small energy could emerge and become clearer,
so that the separation is more accurate.
Compared to previous methods,
MP-Net obtains state-of-the-art results on two large scale datasets,
across mixtures with different types and numbers of sounds.
\end{abstract}
\vspace{-10pt}

%% file: intro.tex
\section{Introduction}
\label{sec:intro}

Besides visual cues,
the sound that comes along with what we see 
often provides complementary information,
which could be used for object detection \cite{girshick2015fast, he2017mask, li2019feature, li2018neural} to clarify ambiguous visual cues,
and description generation \cite{dai2017towards, dai2018rethinking, xiong2018move, dai2018neural} to enrich semantics.
On the other hand, 
as what we hear in most cases is the mixture of different sounds,
coming from different sources,
it is necessary to separate sounds and associate them to sources in the visual scene,
before utilizing sound data.

The difficulties of visual sound separation lie in several aspects.
1) First,
possible sound sources in the corresponding videos may not make any sound,
which causes ambiguity.
2) Second,
the mixture usually contains a large variance in terms of numbers and types.
3) More importantly,
sounds in the mixture often affect each other in multiple ways.
For example,
sounds with large energy often dominate the mixture,
making other sounds less distinguishable or
even sound like noise in some cases.

Existing works \cite{gao2018learning, zhao2018sound} on visual sound separation mainly 
separate each sound independently.
They assume either fixed types or fixed numbers of sounds,
separating sounds independently.
Since strong assumptions in \cite{gao2018learning, zhao2018sound} have limited their applicability in generalized scenarios,
separating sounds independently could lead to inconsistency between the actual mixture and the mixture of separated sounds,
\eg~some data in the actual mixture does not appear in any sounds.
Moreover,
the separation of sounds with small energy may be affected by sounds with large energy in such independent processes.
 
Facing these challenges,
we propose a novel solution,
referred to as MinusPlus Network (MP-Net),
which identifies each sound in the mixture recursively, in descending order of average energy.
It can be divided into two stages,
namely a \emph{minus} stage and a \emph{plus} stage.
At each step of the minus stage,
MP-Net identifies the most salient sound from the current mixture,
then removes the sound therefrom.
This process repeats until the current mixture becomes empty or contains only noise.
Due to the removal of preceding separations,
only one sound could obtain the component that is shared by multiple sounds.
Consequently,
to compensate such cases,
MP-Net refines each sound in the plus stage,
which computes a residual based on the sound itself and the mixture of preceding separated sounds.
The final sound is obtained by mixing the outputs of both stages.
 
MP-Net efficiently overcomes the challenges of visual sound separation.
By recursively separating sounds,
it adaptively decides the number of sounds in the mixture,
without knowing a priori the number and the types of sounds.
Moreover,
in MP-Net,
sounds with large energy will be removed from the mixture after they are separated.
In this way,
sounds with relatively smaller energy naturally emerge and become clearer,
diminishing the effect of imbalanced sound energy.

Overall, our contributions can be briefly summarized as follows:
(1) We propose a novel framework, referred to as MinusPlus Network (MP-Net),
 to separate independent sounds from the recorded mixture based on a corresponding video. 
Unlike previous works which assume a fixed number of sounds in the mixture, 
the proposed framework could dynamically determine the number of sounds, leading to better generalization ability.
(2) MP-Net utilizes a novel way to alleviate the issue of imbalanced energy of sounds in the mixture,
by subtracting salient sounds from the mixture after they are separated,
so that sounds with less energy could emerge.
(3) On two large scale datasets,
MP-Net obtains more accurate results,
and generalizes better compared to the state-of-the-art method.

%% file: relwork.tex
\section{Related Work}

Works connecting visual and audio data can be roughly divided into several categories.

The first category is jointly embedding audio-visual data.
Aytar \etal~\cite{aytar2016soundnet} transfer discriminative knowledge in visual content to audio data by minimizing the KL divergence of their representations.
Arandjelovic \etal~\cite{arandjelovic2017look} associate representations of visual and audio data by learning their correspondence (\ie~whether they belong to the same video),
and authors in \cite{owens2016ambient, owens2018audio, korbar2018cooperative, zhou2018talking} further extend such correspondence to temporal alignment,
resulting in better representations. 
Different from these works,
visual sound separation requires to separate each independent sound from the mixture,
relying on the corresponding video.

The task of sound localization also requires jointly processing visual and audio data,
which identifies the region that generates the sound.
To solve this task,
Hershey \etal~\cite{hershey2000audio} locate sound sources in video frames by measuring audio-visual synchrony.
Both Tian \etal~\cite{tian2018audio} and Parascandolo \etal~\cite{arandjelovic2018objects} apply sound event detection to find sound sources.
Finally, Senocak \etal~\cite{senocak2018learning} and Arandjelovic \etal~\cite{arandjelovic2018objects} find sound sources by analyzing the activation of feature maps.
Although visual sound separation could also locate separated sounds in the corresponding video,
it requires separating the sounds at first,
making it more challenging.

Visual sound separation belongs to the third category,
a special type of which is visual speech separation,
where sounds in the mixture are all human speeches.
For example,
Afouras \etal~\cite{afouras2018conversation} and Ephrat \etal~\cite{ephrat2018looking} obtain a speaker-independent model by leveraging a large amount of news and TV videos,
and Xu \etal~\cite{xu2018modeling} propose an auditory selection framework which uses attention and memory to capture speech characteristics.
Unlike these works,
we target the general task of separating sounds with different types,
which have more diverse sound characteristics.

The most related works are \cite{zhao2018sound} and \cite{gao2018learning}.
In \cite{gao2018learning},
a convolutional network is used to predict the type of objects appeared in the video,
and Non-negative Matrix Factorization \cite{fevotte2009nonnegative}~is used to extract a set of basic components.
The association between each object and each basic component will be estimated via a Multi-Instance Multi-Label objective.
Consequently,
sounds will be separated using the associations between basic components and each predicted object.
\cite{zhao2018sound} follows a similar framework,
replacing Non-negative Matrix Factorization with a U-Net \cite{ronneberger2015u}.
In addition, instead of predicting object-base associations,
it directly predicts weights conditioned on visual semantics.
While the former predicts the existence of different objects in the video,
assuming fixed types of sounds,
the latter assumes a fixed number of sounds.
Such strong assumptions have limited their generalization ability,
as the mixture of sounds often has large variance across sound types and numbers.
More importantly,
each prediction in \cite{zhao2018sound} and \cite{gao2018learning} 
is conducted independently.
As a result,
1) there may be an inconsistency between the mixture of all predicted sounds and the actual mixture.
\eg~some data appeared in the actual mixture may not appear in any predicted sounds,
or some data has appeared too many times in predicted sounds,
exceeding its frequency in the actual mixture.
2) As sounds in the mixture have different average energy,
sounds with large energy may affect the prediction accuracy of sounds with less energy.
Different from them,
our proposed method recursively predicts each sound in the mixture,
following the order of average energy.
The predicted sound with a large energy will be removed from the mixture after its prediction.
In this way,
our proposed method requires no assumptions on the type and number of sounds
and ensures consistent predictions with the input mixture.
Moreover,
when sounds with large energy are removed from the mixture continually,
sounds with less energy could emerge and become clearer,
resulting in more accurate predictions.

%% file: frmwork.tex
\section{Visual Sound Separation}

In the task of visual sound separation,
we are given a context video $V$ and a recorded mixture of sounds $\mS^{\text{mix}}$,
which is the mixture of a set of independent sounds $\{\mS^{\text{solo}}_1, \mS^{\text{solo}}_2, ..., \mS^{\text{solo}}_n\}$.
The objective is to separate each sound from the mixture based on the visual context in $V$.

We propose a new framework for visual sound separation,
referred to as MinusPlus Network (MP-Net),
which learns to separate each independent sound from the recorded mixture,
without knowing a priori the number of sounds in the mixture (\ie~$n$).
In addition,
MP-Net could also associate each independent sound with a plausible source in the corresponding visual content,
providing a way to link data in two different modalities.

\begin{figure*}[!t]
	\centering
	\includegraphics[width=\textwidth]{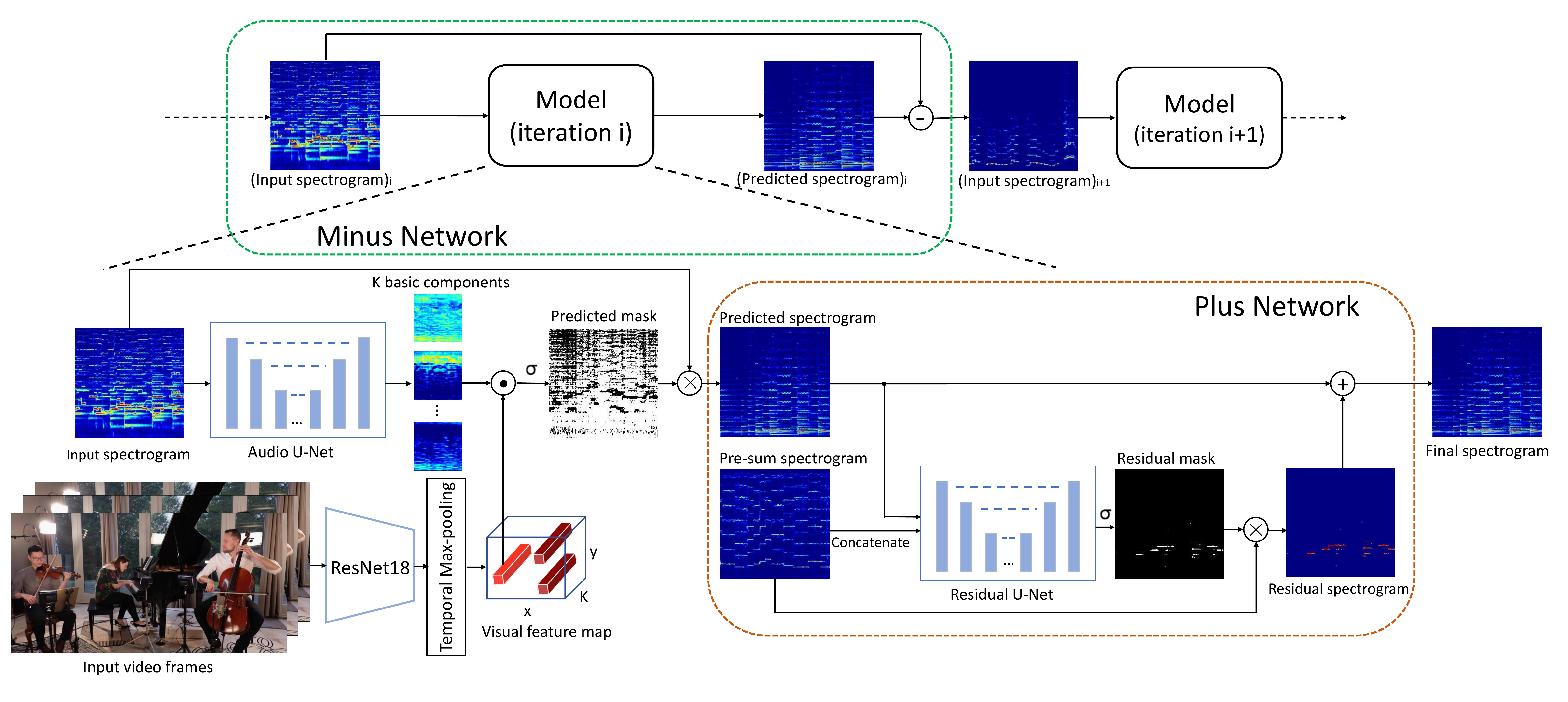}
	\caption{\small The proposed MinusPlus Network (MP-Net) for visual sound separation.
		It consists of two sub-networks, namely the Minus Network (M-Net) and the Plus Network (P-Net).
		In M-Net,
		sounds are recursively separated based on the input video.
		At $i$-th recursive step, a U-Net \cite{ronneberger2015u}~is used to predict $k$ basic components for current mixture,
		which are then used to estimate a mask $\mM$, as well as determine a source in the video for the sound to be separated.
		Based on the mask and visual cues at the sound source,
		a sound is separated, which will be removed from the mixture.
		M-Net repeats these operations until the mixture contains only noise.
		All separated sounds will be refined by P-Net,
		which computes a residual from the mixture of preceding separated sounds.
		The final output of MP-Net for each sound is obtained by mixing the outputs of M-Net and P-Net.
	}
	\label{fig:overview}
\end{figure*}

\subsection{Overview}

In MP-Net, sound data is represented as spectrograms,
and the overall structure of MP-Net has been demonstrated in Figure \ref{fig:overview}.
It has two stages, namely the minus stage and the plus stage.

\textbf{Minus Stage.} 
In the minus stage,
MP-Net recursively separates each independent sound from the mixture $\mS^\text{mix}$,
where at every recursive step it will focus on the sound that is the most salient one in the remaining sounds.
The process could be described as:
\begin{align}
	\mS^\text{mix}_0 & = \mS^\text{mix} \\ 
	\mS^\text{solo}_i & = \text{M-Net}(V, \mS^\text{mix}_{i - 1}) \\
	\mS^\text{mix}_i & = \mS^\text{mix}_{i - 1} \ominus \mS^\text{solo}_i \label{eq:minus}
\end{align}
where $\mS^\text{solo}_i$ is $i$-th predicted sound,
M-Net stands for the sub-net used in the minus stage,
and $\ominus$ is the element-wise subtraction on spectrograms.
As shown in Eq.\eqref{eq:minus},
MP-Net keeps removing $\mS^\text{solo}_i$ from previous mixture $\mS^\text{mix}_{i-1}$,
until current mixture $\mS^\text{mix}_i$ is empty or contains only noise with significantly low energy.

\textbf{Plus Stage.}
While in the minus stage
we remove preceding predictions from the mixture by subtraction,
a prediction $\mS^\text{solo}_i$ may miss some content that is shared by it and preceding predictions $\{\mS^\text{solo}_1, ..., \mS^\text{solo}_{i-1}\}$.
Inspired by this,
MP-Net contains a plus stage,
which further refines each separated sound following:
\begin{align}
	\mS^\text{remix}_i & = \mS^\text{solo}_1 \oplus \cdots \oplus \mS^\text{solo}_{i - 1} \label{eq:remix} \\
	\mS^\text{residual}_i & = \text{P-Net}(\mS^\text{remix}_i, \mS^\text{solo}_i) \\ 
	\mS^\text{solo, final}_i & = \mS^\text{solo}_i \oplus \mS^\text{residual}_i \label{eq:plus}
\end{align}
where P-Net stands for the sub-network used in the plus stage,
and $\oplus$ is the element-wise addition on spectrograms.
As shown in Eq.\eqref{eq:plus},
MP-Net computes a residual $\mS^\text{residual}_i$ for $i$-th prediction,
based on $\mS^\text{solo}_i$ and the mixture of all preceding predictions,
and finally refines $i$-th prediction by mixing $\mS^\text{solo}_i$ and $\mS^\text{residual}_i$.
Subsequently,
in practice,
we use $\mS^\text{solo, final}_i$ instead of $\mS^\text{solo}_i$ in Eq.\eqref{eq:minus} and Eq.\eqref{eq:remix}.

The benefits of using two stages lie in several aspects.
1) The minus stage could effectively determine the number of independent sounds in the mixture,
without knowing it a priori.
2) Removing preceding predictions from the mixture
could diminish their disruption on the remaining sounds,
so that remaining sounds continue emerging as the recursion goes.
3) Removing preceding predictions from the mixture 
potentially helps M-Net focus on the distinct characteristics of remaining sounds,
enhancing the accuracy of its predictions.   
4) The plus stage could compensate for the loss of shared information between each prediction and all its preceding predictions,
potentially smoothing the final prediction of each sound.
Subsequently, we will introduce the two subnets, namely M-Net and P-Net, respectively.

\subsection{M-Net}

M-Net is the sub-net responsible for separating each independent sound from the mixture,
following a recursive procedure.
Specifically,
To separate the most salient sound $\mS^\text{solo}_i$ at $i$-th recursive step,
M-Net will predict $k$ sub-spectrograms $\{\mS^\text{sub}_1, ..., \mS^\text{sub}_k\}$ using a U-Net \cite{ronneberger2015u},
which capture
different patterns in $\mS^\text{mix}$.
At the same time,
we will obtain a feature map $\mV$ of size $H/16 \times W/16 \times k$ from the input video $V$,
which estimates the association score between each sub-spectrogram and the visual content at different spatial location.
With $\mV$ and $\{\mS^\text{sub}_1, \mS^\text{sub}_2, ..., \mS^\text{sub}_k\}$,
we could then identify the associated visual content for $\mS^\text{solo}_i$:
\begin{align}
	(x^\star, y^\star) = \argmax_{(x, y)} \, E \left[ \sigma \Big(\sum_{j=1}^k \mV(x, y, j) * \mS^\text{sub}_j \Big) * \mS^\text{mix} \right], \label{eq:location}
\end{align}
where $\sigma \Big(\sum_{j=1}^k \mV(x, y, j) * \mS^\text{sub}_j \Big)$ computes a location-specific mask.
$E[\cdot]$ computes the average energy of a spectrogram,
and $\sigma$ stands for the sigmoid function,
We regard $(x^\star, y^\star)$ as the source location of $\mS^\text{solo}_i$,
and the feature vector $\vv$ in $\mV$ at that location as the visual feature of $\mS^\text{solo}_i$.
To separate $\mS^\text{solo}_i$,
we reuse the vector $\vv$ as the attention weights on 
sub-spectrograms 
and get the actual mask $\mM$ by 
\begin{align}
 \mM = \sigma(\sum_{j = 1}^k \vv_j \mS^\text{sub}_j),
\end{align}

where $\sigma$ stands for the sigmoid function.
Following \cite{zhao2018sound},
we refer to $\mM$ as a ratio mask,
and an alternative choice is to further binarize $\mM$ to get a binary mask.
Finally,
$\mS^\text{solo}_i$ is obtained by $\mS^\text{solo}_i = \mM \otimes \mS^\text{mix}$.
It is worth noting that we could also directly predict $\mS^\text{solo}_i$ 
following $\mS^\text{solo}_i = \sum_{j = 1}^k \vv_j \mS^\text{sub}_j$.
However,
it is reported that an intermediate mask leads to better results \cite{zhao2018sound}.
At the end of $i$-th recursive step,
MP-Net will remove the predicted $\mS^\text{solo}_i$ from previous mixture $\mS^\text{mix}_{i-1}$ via $\mS^\text{mix}_i = \mS^\text{mix}_{i - 1} \ominus \mS^\text{solo}_i$,
so that less salient sounds could emerge in later recursive steps.
When the average energy in $\mS^\text{mix}_i$ less than a threshold $\epsilon$,
M-Net stops the recursive process,
assuming all sounds have been separated.

\begin{figure}
	\centering
	\includegraphics[width=0.5\textwidth]{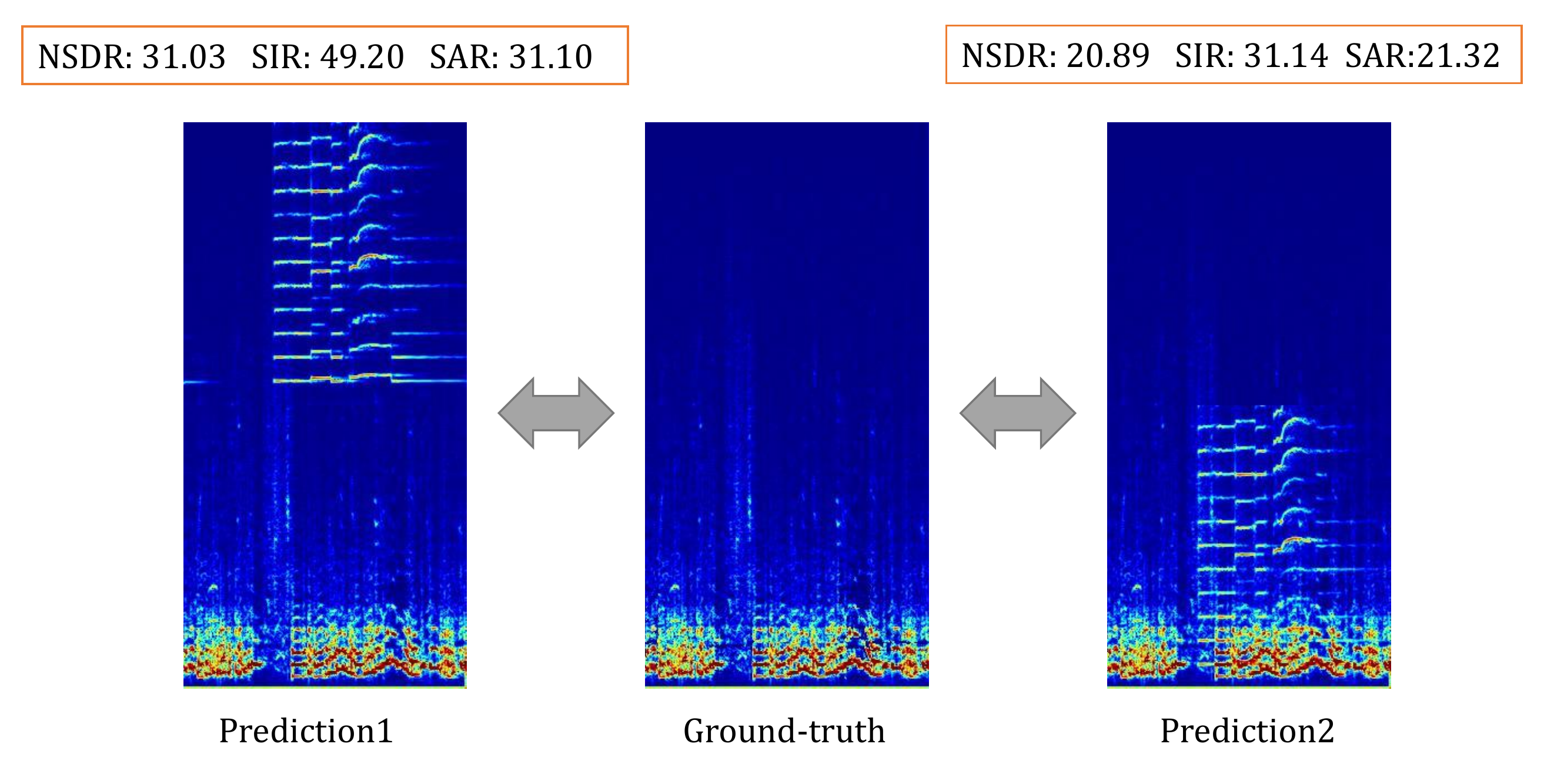}
	\caption{As shown in this figure, when missmatch appears at different locations of spectrograms, scores in terms of SDR, SIR and SAR can vary significantly.}
	\label{fig:metric}
\end{figure}

\subsection{P-Net}

While M-Net makes succeeding predictions more accurate by removing preceding predictions from the mixture,
succeeding predictions may miss some content shared by preceding predictions,
leading to incomplete spectrograms.
To overcome this issue,
MP-Net further applies a P-Net to refine sounds separated by the M-Net.
Specifically,
for $\mS^\text{solo}_i$,
P-Net applies a U-Net \cite{ronneberger2015u} to get a residual mask $\mM_r$,
based on two inputs,
namely $\mS^\text{solo}_i$ and $\mS^\text{remix}_i = \mS^\text{solo}_1 \oplus ... \oplus \mS^\text{solo}_{i-1}$,
which is the re-mixture of preceding predictions,
as the missing content of $\mS^\text{solo}_i$ could only appear in them.
The final sound spectrogram for $i$-th sound is obtained by:
\begin{align}
	\mS^\text{residual}_i & = \mS^\text{remix}_i \otimes \mM_r, \\
	\mS^\text{solo, final}_i & = \mS^\text{solo}_i \oplus \mS^\text{residual}_i.
\end{align}

\subsection{Mutual Distortion Measurement}

To evaluate models for visual sound separation,
previous approaches \cite{gao2018learning, zhao2018sound} utilize 
Normalized Signal-to-Distortion Ratio (NSDR),
Signal-to-Interference Ratio (SIR),
and Signal-to-Artifact Ratio (SAR).
While these traditional metrics could reflect the separation performance to some extent,
they are sensitive to frequencies,
so that scores of different separated sounds are not only affected by their similarities with ground-truths,
but also the locations of mismatches.
Consequently,
as shown in Figure \ref{fig:metric},
scores in terms of SDR, SIR and SAR vary significantly when the mismatch appears at different locations.
To compensate such cases,
we propose to measure the quality of visual sound separation 
under the criterion that two pairs of spectrograms need to obtain approximately the same score if they have the same level of similarities.
This metric,
referred to as Average Mutual Information Distortion (AMID),
computes the average similarity between a separated sound and a ground-truth of another sound,
where the similarity is estimated via the Structural Similarity (SSIM) \cite{wang2004image}~over spectrograms.
Specifically,
for a set of separated sounds $\{\mS^\text{solo}_1, ..., \mS^\text{solo}_m\}$ and its corresponding annotations $\{\mS^\text{gt}_1, ..., \mS^\text{gt}_m\}$,
AMID is computed as:
\begin{align}
	\text{AMID}(\{\mS^\text{solo}_i\}, \{\mS^\text{gt}_j\}) = \frac{1}{m(m-1)} \sum_{i \ne j} \text{SSIM}(\mS^\text{solo}_i, \mS^\text{gt}_j).
\end{align}

As AMID relies on SSIM over spectrograms,
it is insensitive to frequencies. 
Moreover,
a low AMID score indicates the model can distinctly separate sounds in a mixture,
which meets the evaluation requirements of visual sound separation.

%% file: exprt.tex
\section{Experiments}


\input{tab_vegas.tex}

\input{tab_music.tex}

\begin{figure*}[!t]
	\begin{center}
		\includegraphics[width=0.85\textwidth]{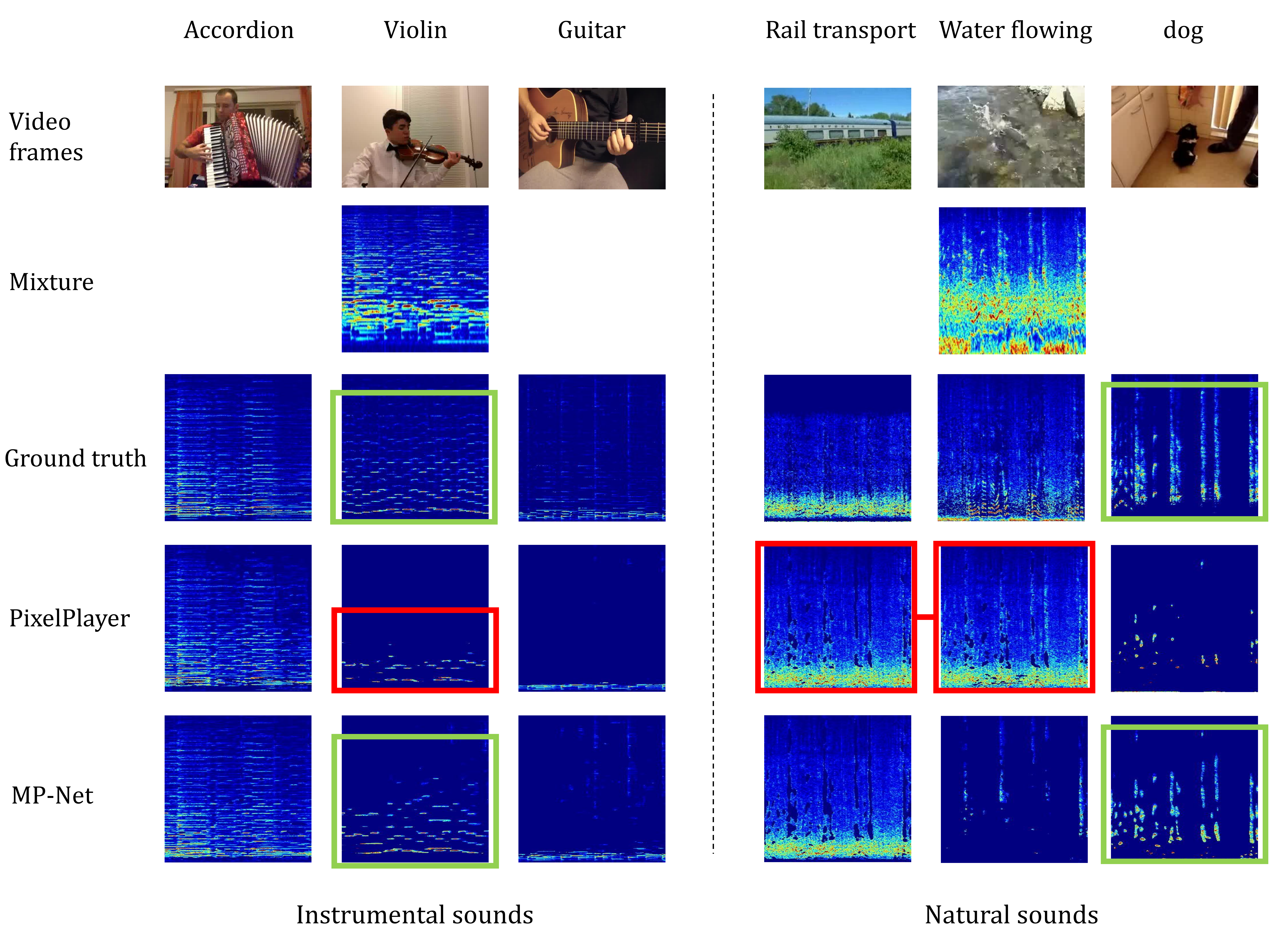}
	\end{center}
	\vspace{-10pt}
	\caption{\small
		Qualitative results for visual sound separation, from MP-Net and PixelPlayer \cite{zhao2018sound}.
		On the left, the mixture of instrumental sounds is demonstrated,
		where MP-Net successfully separates violin's sound, unlike its baseline.
		And on the right, natural sounds are separated from the mixture.
		As the sound of \emph{rail transport} and \emph{water flowing} share a high similarity,
		MP-Net separates dog sounds but predicts silence for \emph{water flowing},
		while its baseline reuses the sound of \emph{rail transport}.
	}

	\label{fig:qualrst}
	\vspace{-5pt}
\end{figure*}

\begin{figure*}[t]
\centering
		\includegraphics[width=0.95\textwidth]{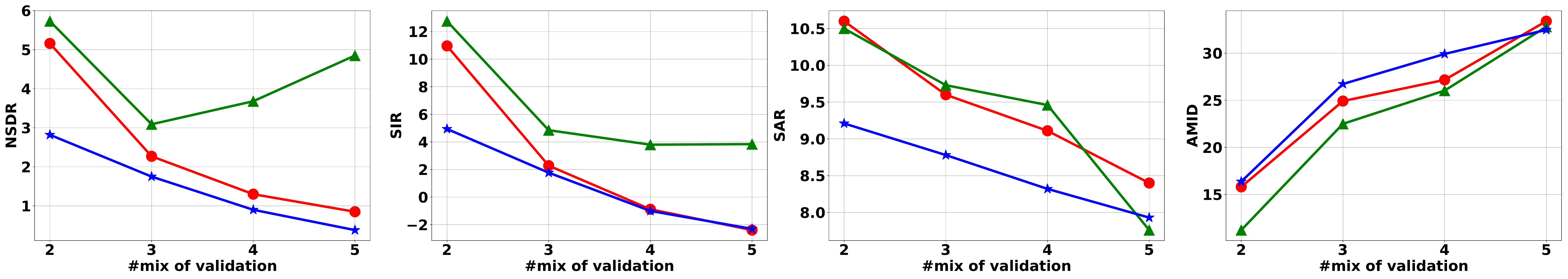}
		\\
		\includegraphics[width=0.95\textwidth]{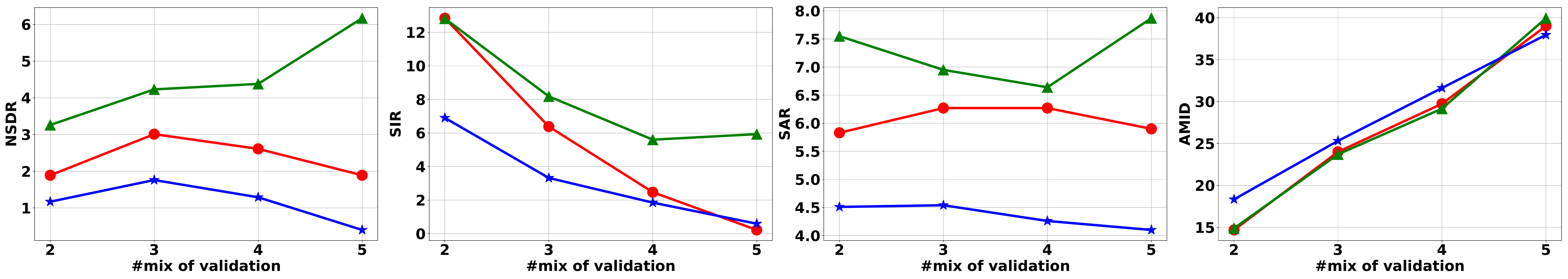}
	\caption{\small
		Curves of varying the number of sounds in the testing mixture on MUSIC, 
		obtained by models respectively trained with mixtures of $2$ sounds (first row), 
		and $3$ sounds (second row). 
		\textcolor{green}{green}, \textcolor{red}{red}, and \textcolor{blue}{blue} lines respectively stand for MP-Net, PixelPlayer \cite{zhao2018sound} and MIML~\cite{gao2018learning}.
	}
	\label{fig:curve_music}
\end{figure*}
\begin{figure*}[t]
	\includegraphics[width=0.95\textwidth]{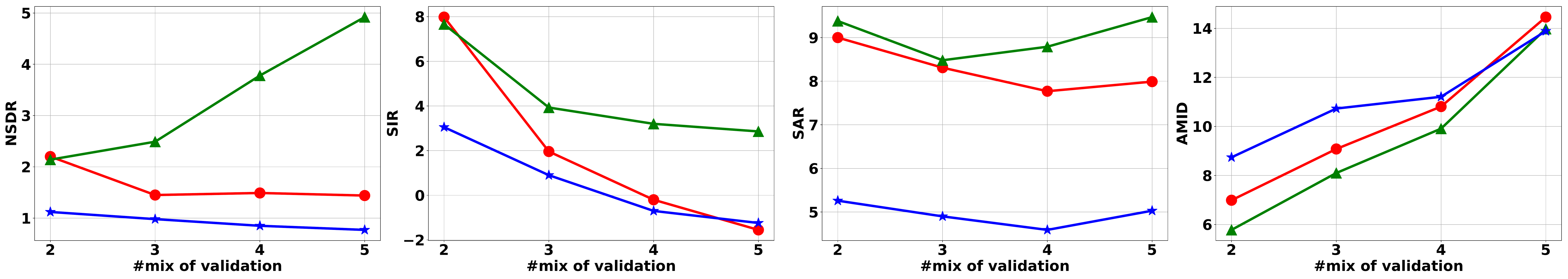}
	\\
	\includegraphics[width=0.95\textwidth]{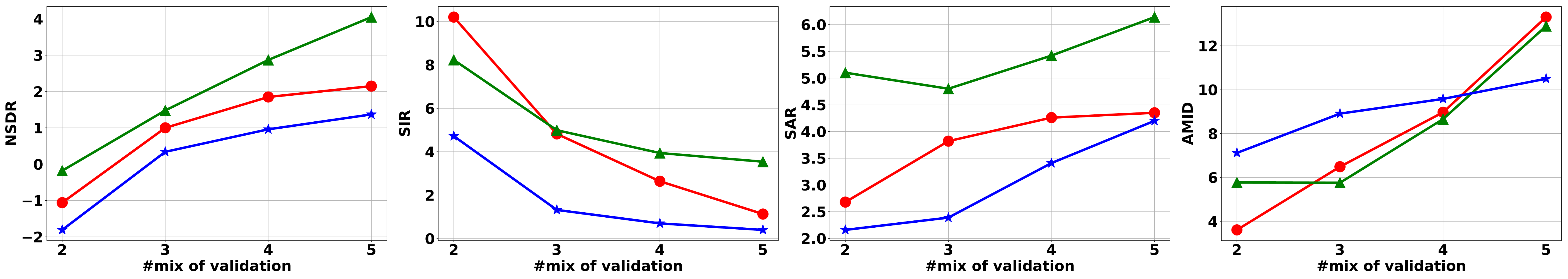}
	\caption{\small
		Curves of varying the number of sounds in the testing mixture on VEGAS,
		obtained by models respectively trained with mixtures of $2$ sounds (first row),
		and $3$ sounds (second row).
		\textcolor{green}{green}, \textcolor{red}{red}, and \textcolor{blue}{blue} lines respectively stand for MP-Net, PixelPlayer \cite{zhao2018sound} and MIML~\cite{gao2018learning}.
	}
	\label{fig:curve_vegas}
	\vspace{-10pt}
\end{figure*}

\begin{figure*}[!t]
	\begin{center}
		\includegraphics[width=0.85\textwidth]{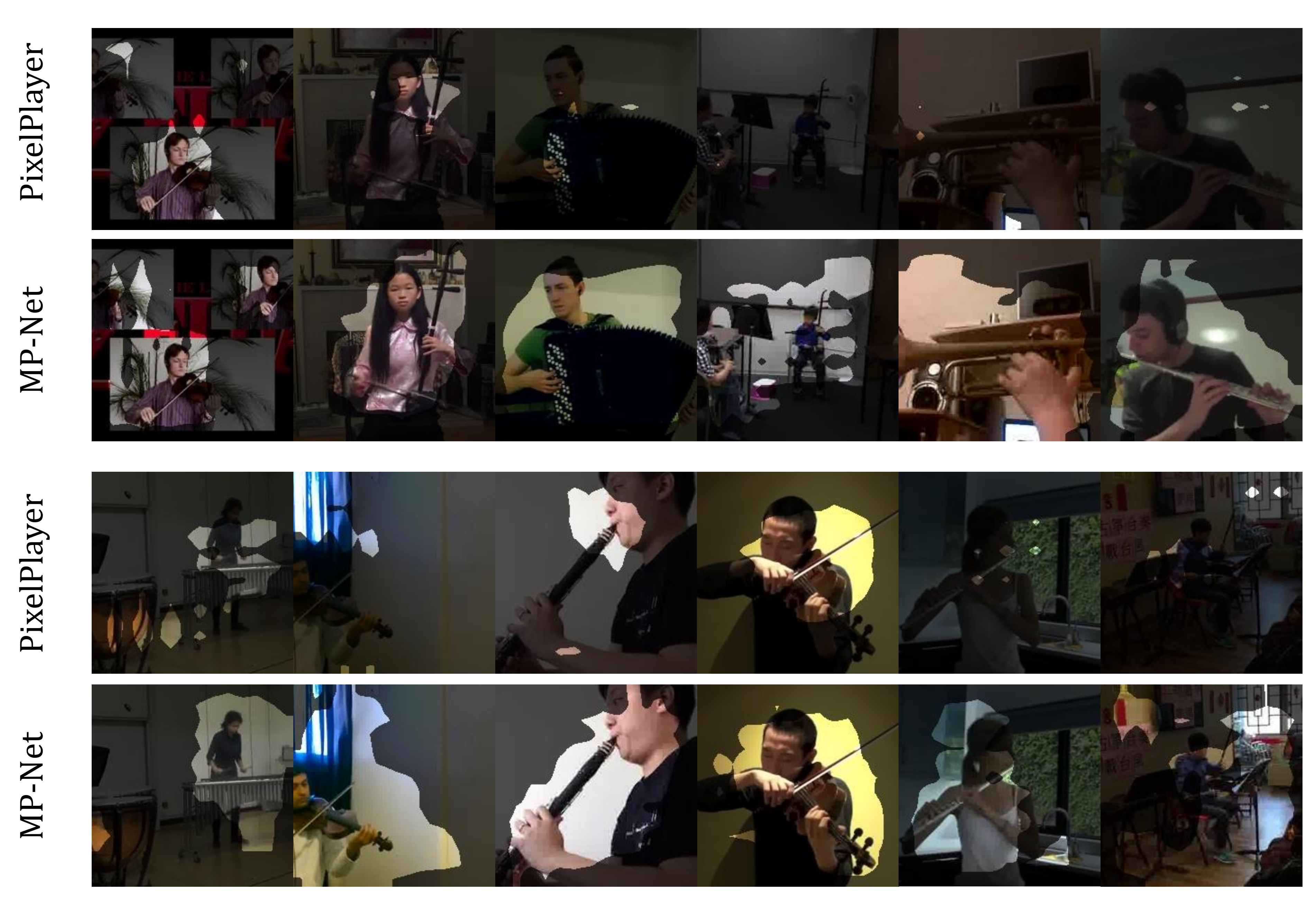}
	\end{center}
	\vspace{-10pt}
	\caption{\small
		Qualitative samples on found associations between visual sound sources (light regions) and different types of sounds,
		using MP-Net and PixelPlayer \cite{zhao2018sound}.
	}
	\label{fig:localization}
	\vspace{-10pt}
\end{figure*}



\subsection{Datasets}

We test MP-Net on two datasets, 
namely Multimodal Sources of Instrument Combinations (MUSIC) \cite{zhao2018sound} 
and Visually Engaged and Grounded AudioSet (VEGAS) \cite{zhou2018visual}.

MUSIC mainly contains untrimmed videos of people playing instruments belonging to $11$ categories,
namely \emph{accordion, acoustic guitar, cello, clarinet, erhu, flute, saxophone, trumpet, tuba, violin} and \emph{xylophone}. 
There are respectively $500$, $130$ and $40$ samples in the train, validation and test set of MUSIC.
While the test set of MUSIC contains only duets without ground-truths of sounds in mixtures,
we use its validation set as test set,
and train set for training and validation.
While MUSIC focuses on instrumental sounds,
VEGAS, another dataset will a larger scale,
covers $10$ types of natural sounds,
including \emph{baby crying, chainsaw, dog, drum, fireworks, helicopter, printer, rail transport, snoring} and \emph{water flowing},
trimmed from AudioSet \cite{gemmeke2017audio}.
$2,000$ samples in VEGAS are used as the test,
with remaining samples being used for training and validation. 

\subsection{Training and Testing Details}

Due to the lack of ground-truth of real mixed data, \ie~those videos that contain multiple sounds.
We construct such data from \emph{solo} video clips instead.
Each clip contains at most one sound.
We denote the collection of solo video clips by $\{\mS^{\text{solo}}_j, V_j\}_{j = 1}^N$, where 
$\mS^{\text{solo}}_j$ and $V_j$ respectively represent the sound and visual content. 
Note that a video clip can be silent, for such case, 
$\mS^{\text{solo}}_j$ is an empty spectrogram. 
For each $V_j$, we sample $T=6$ frames at even intervals, and extract 
visual features for each frame using ResNet-18 \cite{he2016deep}. This would result in 
a feature tensor of size $T \times (H / 16) \times (W / 16) \times k$.
In both training and testing, this feature tensor will be reduced into 
a vector to represent the visual content by performing 
max pooling along the first three dimensions. 
On top of this solo video collection, we then follow the Mix-and-Separate 
strategy as in \cite{zhao2018sound, gao2018learning} to construct the mixed video/sound data, where each 
sample mixes $n$ videos, called a mix-$n$ sample. 

Audios are preprocessed before training and testing.
Specifically,
we sample audios at $16$kHz,
and use the open-sourced package \emph{librosa} \cite{mcfee2015librosa}~to transform the sound clips of around 6 seconds into STFT spectrograms of size $750 \times 256$,
where the window size and the hop length are respectively set as $1,500$ and $375$. We down-sample it on mel scale and obtain the spectrogram with size $256 \times 256$.

We use $k=16$ for the M-Net.
We adopt a three-round training strategy,
where in the first round, we train the M-Net in isolation.
And in the second round, we train the P-Net while fixing parameters of the M-Net.
Finally, in the third round, the M-Net and the P-Net are jointly finetuned.

During training, for each mix-$n$ sample, 
we first perform data augmentation, 
randomly scaling the energy of the spectrograms.
Then, MP-Net makes $n$ predictions in the descending order of the average energy 
of the ground-truth sounds. 
Particularly, for the $t$-th prediction, MP-Net predicts $\mM$ and $\mM_r$
for the sound with $t$-th largest average energy and computes the BCE loss 
between $\mM + \mM_r$ and the ground-truth mask if binary masks are used, 
or $\cL_1$ loss if ratio masks are used. 
After all $n$ predictions are done, we add an extra loss between the 
remaining mixture and an empty spectrogram -- ideally if all $n$ predictions
are precise, there should be no sound left. 

During evaluation, we determine the prediction order by Eq.(7).
Since all baselines need to know the number of sounds in the mixture,
to compare fairly, we also provide the number of sounds to MP-Net.
It is, however, noteworthy that
MP-Net could work without this information,
relying only on the termination criterion to determine the number.
On \emph{MUSIC}, MP-Net predicts the correct number of sounds with over $90\%$ of accuracy.

\subsection{Experimental Results}

\paragraph{Results on Effectiveness}
To study the effectiveness of our model,
we compared our model to state-of-the-art methods,
namely PixelPlayer \cite{zhao2018sound} and MIML \cite{gao2018learning},
across datasets and settings,
providing a comprehensive comparison.
Specifically,
on both MUSIC and VEGAS,
we train and evaluate all methods twice, respectively using mix-$2$ and mix-$3$ samples,
which contain $2$ and $3$ sounds in the mixture.
For PixelPlayer and MP-Net,
we further alter the form of masks to switch between ratio masks and binary masks.
The results in terms of NSDR, SIR, SAR and AMID are listed in Table \ref{tab:vegas} for VEGAS and Table \ref{tab:music} for MUSIC.
We observe that
1) our proposed MP-Net obtains best results in most settings,
outperforming PixelPlayer and MIML by large margins,
which indicate the effectiveness of separating sounds in the order of average energy.
2) Using ratio masks is better in terms of NSDR and SAR,
while using binary masks is better in terms of SIR and AMID.
3) Our proposed metric AMID correlates well with other metrics,
which intuitively verifies its effectiveness.
4) Scores of all methods on mix-$2$ samples are much higher than scores on mix-$3$ samples,
which add only one more sound in the mixture.
Such differences in scores have shown the challenges of visual sound separation.
5) In general,
methods obtain higher scores on MUSIC, 
meaning natural sounds are more complicated than instrumental sounds,
as instrumental sounds often contain regular patterns.

\vspace{-8pt}
\paragraph{Results on Ablation Study}
While the proposed MP-Net contains two sub-nets,
we have compared MP-Net with and without P-Net.
As shown in Table \ref{tab:vegas} and Table \ref{tab:music},
on all metrics,
MP-Net with P-Net outperforms MP-Net without P-Net by large margins,
indicating
1) different sounds have shared patterns,
a good model needs to take this into consideration,
so that the mixture of separated sounds is consistent with the actual mixture.
2) P-Net could effectively compensate the loss of shared patterns caused by sound removements,
filling blanks in the spectrograms.

\vspace{-8pt}
\paragraph{Results on Robustness}
A benefit of recursively separating sounds from the mixture 
is that MP-Net is robust when the number of sounds in the mixture varies,
although trained with a fixed number of sounds.
To verify the generalization ability of MP-Net,
we have tested all methods that trained with mix-$2$ or mix-$3$ samples,
on samples with an increasing number of sounds in the mixtures.
The resulting curves on MUSIC are shown in Figure \ref{fig:curve_music},
and Figure \ref{fig:curve_vegas}~includes the curves on VEGAS.
In Figure \ref{fig:curve_music}~and Figure \ref{fig:curve_vegas},
up to mixtures consisting of $5$ sounds,
MP-Net trained with a fixed number of sounds in the mixture
outperforms baselines steadily as the number of sounds in the mixture increases.

\vspace{-8pt}
\paragraph{Qualitative Results}
In Figure \ref{fig:qualrst},
we show qualitative samples with sounds separated by respectively MP-Net and PixelPlayer,
in the form of spectrograms.
In the sample with a mixture of instrumental sounds,
PixelPlayer fails to separate sounds belonging to violin and guitar,
as their sounds are overwhelmed by the sound of accordion.
On the contrary,
MP-Net successfully separates sounds of violin and guitar,
alleviating the effect of the accordion's sound.
Unlike PixelPlayer that separates sounds independently,
MP-Net recursively separates the dominant sound in current mixture,
and removes it from the mixture,
leading to accurate separation results.
A similar phenomenon can also be observed 
in the sample with a mixture of natural sounds,
PixelPlayer predicts the same sound for \texttt{rail\_transport} and \texttt{water\_flowing},
and fails to separate the sound of a dog.



\vspace{-8pt}
\paragraph{Localization Results}
MP-Net could also be used to associate sound sources in the video with separated sounds,
using Eq.\eqref{eq:location}.
We show some samples in Figure \ref{fig:localization},
where compared to PixelPlayer,
MP-Net produces more precise associations between separated sounds and their possible sound sources.

%% file: tab_vegas.tex
\begin{table*}[!t]
	\centering
	\small
	\begin{tabular}{cc cccc cccc} 
		\toprule
		&& \multicolumn{4}{c}{mix-$2$} & \multicolumn{4}{c}{mix-$3$} \\
		\cmidrule(lr){3-6} \cmidrule(lr){7-10} 
              & Mask & NSDR$\uparrow$ & SIR$\uparrow$ & SAR$\uparrow$ & AMID$\downarrow$  & NSDR$\uparrow$ & SIR$\uparrow$ & SAR$\uparrow$ & AMID$\downarrow$ \\           	
		\midrule
		MIML \cite{gao2018learning} & - & 1.12 & 3.05 & 5.26 & 8.74 & 0.34 & 1.32 & 2.39 & 8.91 \\
		\midrule	
	\multirow{2}{*}{PixelPlayer\cite{zhao2018sound}} & Binary & 2.20 & \textbf{7.98} & 9.00 & 6.99 & 1.00 & 4.82 & 3.82 & 6.49 \\
							& Ratio & \textbf{2.96} & 5.91 & 13.77 & 10.35 & 2.99 & 2.59 & 10.69 & 10.55 \\
		\midrule
	\multirow{2}{*}{M-Net} & Binary & 2.02 & 7.48 & 9.22 & 5.96 & 1.23 & 4.76 & 4.69 & 5.96 \\
				& Ratio & 2.66 & 5.17 & 14.19 & 6.80 & 3.54 & 2.31 & 15.92 & 11.54 \\
		\midrule
	\multirow{2}{*}{MP-Net (M-Net + P-Net)} & Binary & 2.14 & 7.66 & 9.47 & \textbf{5.78} & 1.48 & \textbf{4.99} & 4.80 & \textbf{5.76} \\ 
				& Ratio & 2.81 & 5.45 & \textbf{14.49} & 6.53 & \textbf{3.75} &  2.52 & \textbf{16.77} & 10.59 \\
		\bottomrule
	\end{tabular}
	\caption{\small
		This table lists the results of visual sound separation on VEGAS \cite{zhou2018visual},
		where MP-Net obtains best performance under various metrics and settings.
	}
	\label{tab:vegas}
\end{table*}

%% file: tab_music.tex
\begin{table*}[!t]
	\centering
	\small
	\begin{tabular}{cc cccc cccc} 
		\toprule
		&& \multicolumn{4}{c}{mix-$2$} & \multicolumn{4}{c}{mix-$3$} \\ 
		\cmidrule(lr){3-6} \cmidrule(lr){7-10} 
		& Mask & NSDR$\uparrow$ & SIR$\uparrow$ & SAR$\uparrow$ & AMID$\downarrow$  & NSDR$\uparrow$ & SIR$\uparrow$ & SAR$\uparrow$ & AMID$\downarrow$ \\ 
		\midrule
	MIML \cite{gao2018learning} & - & 2.82 & 4.94 & 9.21 & 16.37 & 1.76 & 3.32 & 4.54 & 25.32 \\
		\midrule
\multirow{2}{*}{PixelPlayer\cite{zhao2018sound}} & Binary & 5.16 & 10.96 & 10.60 & 15.81 & 3.01 & 6.38 & 6.27 & 24.01 \\ 
				& Ratio & 6.09 & 8.07 & 14.93 & 18.81 & 4.83 & 4.87 & 11.19 & 29.84 \\ 
\midrule
	\multirow{2}{*}{M-Net} & Binary & 5.47 & 12.63 & 10.21 & 11.83 & 4.01 & 7.89 & 6.76 & 23.76 \\ 
				& Ratio & 6.82 & 10.12 & 14.98 & 13.90 & 5.61 & 5.03 & 13.42 & 24.05 \\ 
		\midrule
	\multirow{2}{*}{MP-Net (M-Net + P-Net)} & Binary & 5.73 & \textbf{12.75} & 10.50 & \textbf{11.22} & 4.23 & \textbf{8.18} & 6.95 & \textbf{23.10} \\ 
				& Ratio & \textbf{7.00} & 10.39 & \textbf{15.31} & 13.36 & \textbf{5.75} & 5.37 & \textbf{13.68} & 23.51 \\ 
		\bottomrule
	\end{tabular}
	\caption{\small
		This table lists the results of visual sound separation on MUSIC \cite{zhao2018sound},
		where MP-Net obtains best performance under various metrics and settings.
	}
	\label{tab:music}
\end{table*}

%% file: concls.tex
\section{Conclusion}
We propose MinusPlus Network (MP-Net), a novel framework 
for visual sound separation.
Unlike previous methods that separate each sound independently,
MP-Net jointly considers all sounds,
where sounds with larger energy are separated firstly,
followed by them being removed from the mixture,
so that sounds with smaller energy keep emerging.
In this way,
once trained, MP-Net could deal with mixtures made of an arbitrary number of sounds.
On two datasets,
MP-Net is shown to consistently outperform state-of-the-arts,
and maintains steady performance as the number of sounds in mixtures increases.
Besides,
MP-Net could also associate separated sounds to possible sound sources in the corresponding video,
potentially linking data from two modalities.

\vspace{-12pt}
\paragraph{Acknowledgement}
This work is partially supported by the Collaborative Research Grant from SenseTime (CUHK Agreement No.TS1610626 \& No.TS1712093),
and the General Research Fund (GRF) of Hong Kong (No.14236516 \& No.14203518).